%% file: neurips_2026.tex
\documentclass{article}

\PassOptionsToPackage{numbers, compress}{natbib}
 \usepackage[preprint]{neurips_2026}

\usepackage[utf8]{inputenc} 
\usepackage[T1]{fontenc}    
\usepackage{hyperref}       
\usepackage{url}            
\usepackage{booktabs}       
\usepackage{amsfonts}       
\usepackage{amsmath}        
\usepackage{nicefrac}       
\usepackage{microtype}      
\usepackage[table]{xcolor}  
\usepackage{multirow}
\usepackage{makecell}
\usepackage{graphicx}
\usepackage{adjustbox} 
\usepackage{array}
\usepackage{capt-of}
\usepackage{stfloats}

\definecolor{oursrowbg}{RGB}{232,242,252}

\title{OmniFocus: Query-Guided Modality-Balanced Token Compression for Omni-Modal Large Language Models}

\author{
  \textbf{Shijie Cao}\textsuperscript{1,2},
  \textbf{Qingyu Zhang}\textsuperscript{2},
  \textbf{Boxi Yu}\textsuperscript{3},
  \textbf{Yuzhong Zhang}\textsuperscript{4},\\
  \textbf{Boxi Cao}\textsuperscript{2},
  \textbf{Yaojie Lu}\textsuperscript{2},
  \textbf{Hongyu Lin}\textsuperscript{2},
  \textbf{Xianpei Han}\textsuperscript{2},
  \textbf{Le Sun}\textsuperscript{2},
\\
  \textsuperscript{1}School of Advanced Interdisciplinary Sciences, \\University of Chinese Academy of Sciences,\\
  \textsuperscript{2}Chinese Information Processing Laboratory, Institute of Software,\\ Chinese Academy of Sciences \\
  \textsuperscript{3}University of Limerick
  \textsuperscript{4}CUHK, Shenzhen \\
  \texttt{\{boxi2020, luyaojie, hongyu, xianpei, sunle\}@iscas.ac.cn}\\
  \texttt{caoshijie22@mails.ucas.ac.cn} \texttt{boxiyu@link.cuhk.edu.cn}\\
  \texttt{123090848@link.cuhk.edu.cn}
}

\begin{document}

\maketitle

\begin{abstract}
Omni modal large language models (OmniLLMs) have attracted wide attention for their ability to jointly 
process audio and video, but they generate large token sequences under audio-visual inputs, leading to
substantial inference cost.
Existing audio-visual token compression methods often rely on unimodal guidance, overlooking the temporal
locality of query-relevant evidence in audio-visual inputs and implicitly assuming that the two modalities 
share a temporally aligned information density distribution.
We propose \textbf{OmniFocus}, a training-free query-guided token compression method for OmniLLMs that performs independent importance estimation for video and audio, enabling a modality-symmetric compression design that preserves modality-specific salient evidence while maintaining audio-visual alignment, thereby mitigating the modality bias issue that can arise from unimodal-guided compression.
Experiments on the Qwen2.5-Omni model family across four audio-visual benchmarks show that OmniFocus maintains strong compressed performance at low token retention ratios and outperforms existing baselines on several major benchmark scores at 25\% token retention.
On DailyOmni with Qwen2.5-Omni-7B at 25\% token retention, OmniFocus maintains 59.40 accuracy while delivering up to 1.38$\times$ prefill speedup relative to the full-token
baseline, highlighting a favorable practical accuracy-efficiency trade-off.

\end{abstract}

\section{Introduction}
Omni-modal large language models (OmniLLMs) extend language models to joint audio-visual understanding, 
showing broad potential across tasks such as audio-visual understanding~\cite{li2025omnivideobench}, 
social and empathetic understanding~\cite{qin2026humansense,xie2026socialomni,tian2026emoomni}, and audio captioning~\cite{xu2025qwen3,chen2025avocado}.
However, audio-visual inputs generate large token sequences, which limits
the applicability of OmniLLMs in long-form audio-visual settings and resource-constrained environments.   

Token compression as a method to reduce inference cost has been widely used in vision, audio, and long-text settings
~\cite{yang2025visionzip,shang2025llava,tao2025dycoke,huang2025prunevid,lin2025speechprune,shao2025tokens}.
By removing redundant tokens while preserving key evidence, token compression can improve inference 
efficiency and reduce GPU memory usage without sacrificing much performance.
Unlike unimodal compression, token compression in OmniLLMs must preserve audio evidence, visual evidence, and cross-modal alignment simultaneously, 
which poses unique challenges for compression design.

\begin{figure}[t]
    \centering
    \includegraphics[width=\linewidth]{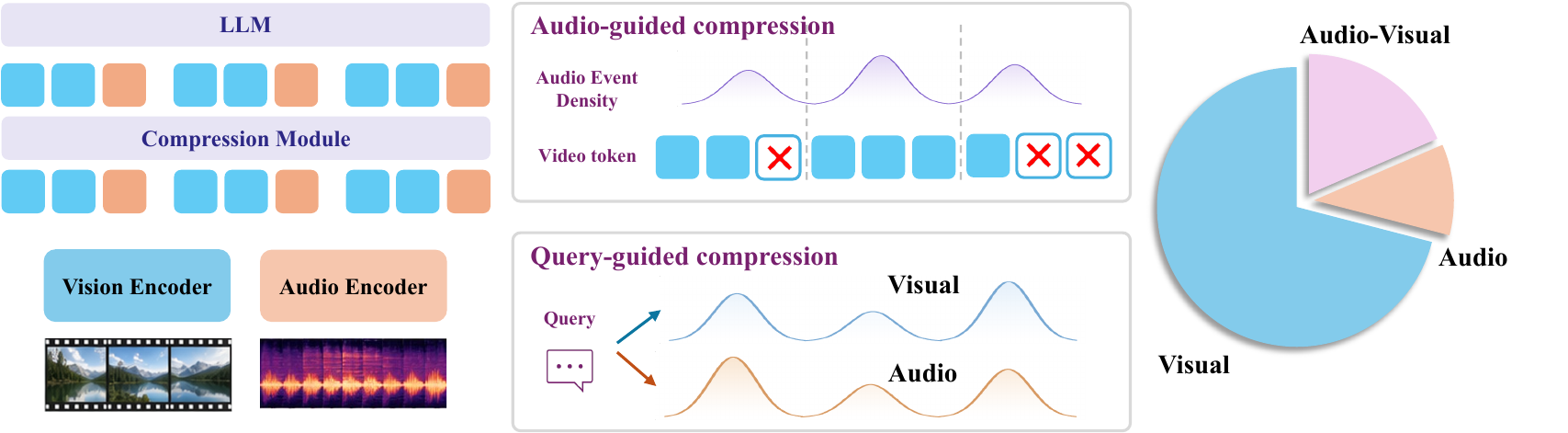}
    \caption{Illustration of the limitation of unimodal-guided audio-visual token compression. Audio-guided compression preserves audio-centric performance relatively well, but degrades more noticeably on visual and joint audio-visual understanding tasks, motivating a query-guided and modality-balanced compression strategy. The modality-type labeling protocol is described in Appendix~\ref{app:modality_type_analysis}.}
    \label{fig:intro}
\end{figure}

In previous work, researchers typically use one modality as a guiding modality to assess the density of 
event information, and design asymmetric compression strategies for the two modalities based on this signal~\cite{tao2025omnizip,ding2026omnisift,li2026dash}.
This unimodal-guided asymmetric compression strategy has two limitations: (1) it overlooks the temporal locality
of query-relevant evidence, and (2) it implicitly assumes that the two modalities share a temporally aligned information
density distribution. 
As illustrated in Figure~\ref{fig:intro}, audio-guided compression preserves audio-centric performance relatively well,
but degrades more noticeably on visual and joint audio-visual understanding tasks. This suggests that using a single 
modality as the compression guide can lead to overfitting to evidence in that modality while overlooking potentially 
important information in the other modality.  

%
To bridge this gap, we propose \textbf{OmniFocus}\footnote{\url{https://github.com/icip-cas/OmniFocus}}, a training-free query-guided token compression method for OmniLLMs.
OmniFocus assesses the importance of different temporal chunks by computing query-token similarity, addressing
the previous methods' neglect of temporal locality and giving higher retention to important temporal chunks.
Performing separate importance estimation for video and audio allows OmniFocus to be a symmetric compression method
that avoids over-reliance on one modality and mitigates the modality bias issue that can arise from unimodal-guided
compression.

Specifically, OmniFocus computes modality-specific chunk importance from
query-token similarity, converts these scores into chunk-wise local drop ratios,
and selects retained tokens using complementary inter-modal association and
intra-modal peak scores. This design preserves both audio-visual correspondence
and modality-specific salient evidence in compressed settings.

%
Experiments on the Qwen2.5-Omni model family across four audio-visual
benchmarks show that OmniFocus achieves a favorable accuracy-efficiency
trade-off and reaches best or tied-best compressed results on several benchmark
scores. These results suggest that query-guided modality-symmetric compression
is a more reliable design choice for long-form audio-visual understanding than
single-modality-guided token budgeting.
For example, on DailyOmni with Qwen2.5-Omni-7B at 25\% token retention, 
OmniFocus achieves 59.40 accuracy while delivering 1.38$\times$ prefill speedup 
and 1.32$\times$ end-to-end speedup relative to the full-token baseline.

%
Our contributions are summarized as follows:
\begin{itemize}
    \item We identify and empirically validate the modality-bias limitation of unimodal-guided audio-visual token compression in OmniLLMs.
    \item We propose OmniFocus, a training-free query-guided token compression method that performs modality-symmetric compression and preserves both inter-modal association and
  intra-modal peak evidence under low token retention ratios.
    \item Extensive experiments on the Qwen2.5-Omni model family across multiple audio-visual benchmarks show that OmniFocus achieves a favorable accuracy-efficiency trade-off and reaches best or tied-best compressed results on several benchmark scores.
\end{itemize}

\section{Related Works}

\subsection{Omni-Modal Large Language Models}
Omni-modal large language models (OmniLLMs) unify visual, auditory, and textual
understanding within a single backbone~\cite{jiang2025specific,xie2024mini,xie2024mini2,xu2025qwen3,team2026qwen3,li2026omnigaia,tian2026emoomni,guan2026thinkomni,yang2025omni}.
Representative proprietary and open-source systems, including GPT-4o,
Gemini~\cite{comanici2025gemini},
Qwen-Omni~\cite{xu2025qwen3,team2026qwen3},
Baichuan-Omni~\cite{li2025baichuan},
Mini-Omni~\cite{xie2024mini,xie2024mini2}, and
InteractiveOmni~\cite{tong2025interactiveomni}, show that joint audio-visual
reasoning can be integrated into a general multimodal model.
Although these systems share the goal of unified audio-visual reasoning, they
differ in how audio and video representations are organized before entering the
language backbone.

Some OmniLLMs organize audio and video tokens into temporally aligned windows
to strengthen cross-modal correspondence, while others use more direct
audio-visual fusion before the language backbone. Despite these design
differences, long audio-visual inputs still produce large multimodal token
sequences and substantially increase inference cost. Recent benchmarks further
require long-form audio-visual reasoning in daily-life videos~\cite{zhou2025daily},
real-world synchronized omnimodal perception~\cite{hong2025worldsense},
synergistic audio-visual reasoning~\cite{li2025omnivideobench}, and video
question answering across different time scales~\cite{fu2025video,goel2026mmou,geng2025longvale}.
These demands make efficient yet modality-faithful audio-visual processing an
important problem.

\subsection{Token Compression for Multimodal Models}
Token compression reduces inference cost by removing redundant tokens while
preserving task-relevant information. Prior work has explored visual token
reduction~\cite{bolya2022token,chen2024image,he2024zipvl,ye2025fit,yang2025visionzip,huang2025prunevid,shao2025holitom,yang2025topv},
speech or audio token pruning~\cite{lin2025speechprune,lee2025token}, and multimodal token
compression for large multimodal models~\cite{shang2025llava,tao2025dycoke,gong2025echoingpixels},
enabling efficient inference without retraining the backbone. Related
efficiency work on long-video understanding also reduces temporal redundancy
through adaptive visual selection or event-focused compression~\cite{tang2025adaptive,chen2026event,di2025streaming}.

Compared with unimodal or vision-language settings, token compression in
OmniLLMs must preserve audio evidence, visual evidence, and cross-modal
alignment simultaneously. OmniZip~\cite{tao2025omnizip} is the closest prior
method in this setting: it estimates temporal importance from audio-side
attention and uses that signal to guide audio-visual token budgets. This
mechanism is efficient, but it can introduce modality bias when query-relevant
evidence is sparse, modality-specific, or temporally misaligned across audio
and video. OmniFocus differs by estimating query relevance separately for audio
and video and performing modality-symmetric compression, rather than using one
modality as the compression guide for the other.

\section{Method}

In this section, we first introduce the audio-video token organization in
OmniLLMs, and then present the overall pipeline of OmniFocus, including
query-aware modality scoring, chunk-wise budget allocation, and token
selection.

\subsection{Preliminary}

Given an audio-visual input $[V, A]$, where $V$ denotes the video stream and $A$
denotes the audio stream, an OmniLLM first maps the two modalities into token
sequences using modality-specific encoders:
\begin{equation}
    x_{a} = \text{Encoder}_{a}(A), \quad x_{v} = \text{Encoder}_{v}(V)
\end{equation}
where $x_{a} \in \mathbb{R}^{N_{a} \times D}$ and $x_{v} \in \mathbb{R}^{N_{v} \times D}$, 
with $N_{a}$ and $N_{v}$ denoting the number of audio and video tokens, respectively. 
The model partitions both streams into aligned temporal windows and arranges the
tokens from the same window sequentially. This yields an interleaved token
sequence
$X_{\mathrm{av}} = [x_{v_1}, x_{a_1}, x_{v_2}, x_{a_2}, \dots, x_{v_T}, x_{a_T}]$,
where $x_{v_i}$ and $x_{a_i}$ denote the video and audio tokens in temporal
chunk $i$. OmniFocus operates on $X_{\mathrm{av}}$ before the tokens are
consumed by the LLM backbone.

\subsection{OmniFocus}

\begin{figure}[htbp]
    \centering
    \includegraphics[width=0.98\linewidth]{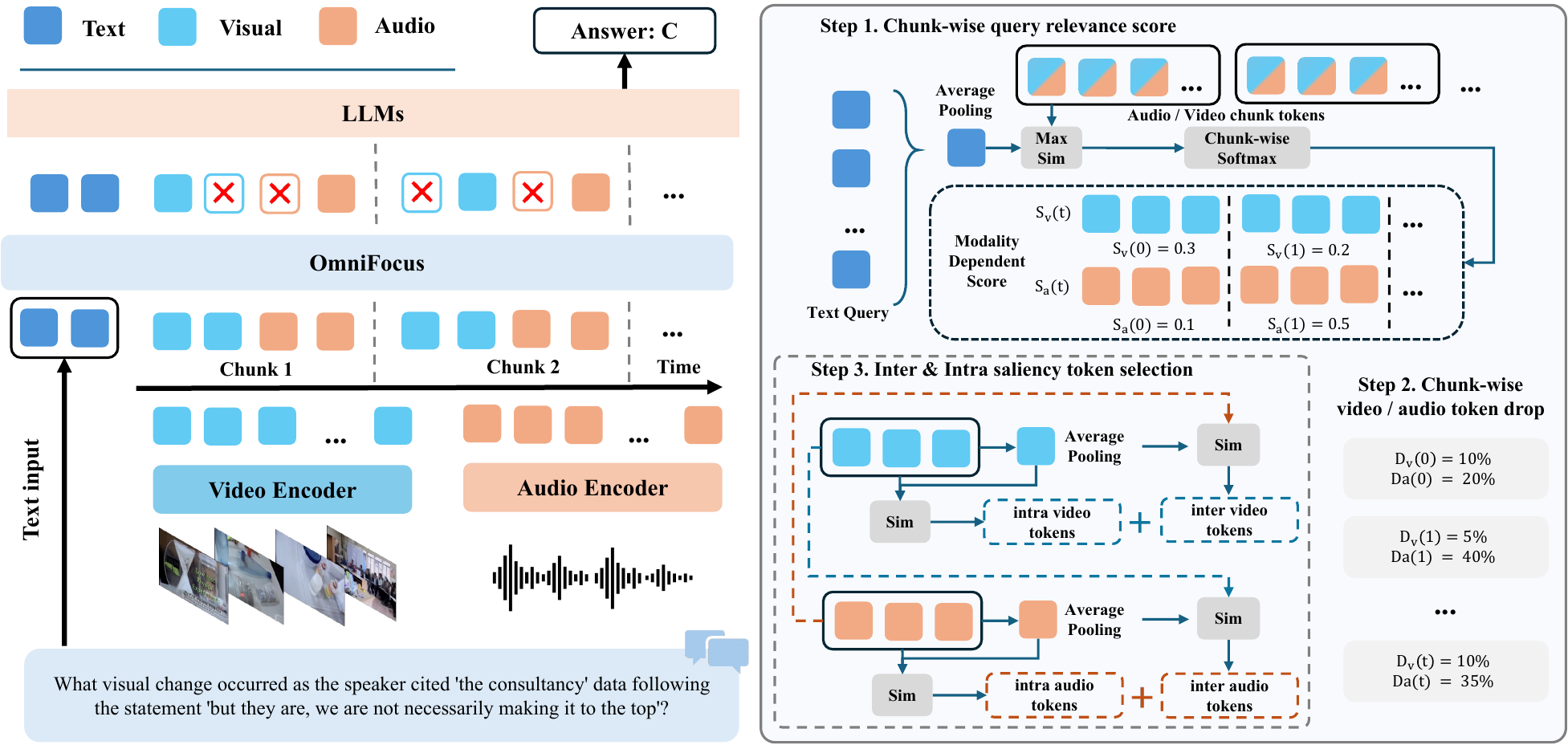}
    \caption{Overview of OmniFocus. Given a textual query and temporally aligned audio-video chunks, OmniFocus estimates modality-specific query relevance, allocates chunk-wise local drop ratios under calibrated global modality budgets, and selects retained tokens using inter-modal association and intra-modal peak scores before feeding the compressed sequence into the LLM backbone.}
    \label{fig:omnifocus}
\end{figure}

Given a textual query $q$ and the interleaved audio-visual token sequence
$X_{\mathrm{av}}$ organized into $T$ temporal chunks, OmniFocus compresses audio
and video tokens before they are fed into the LLM backbone. The method is
training-free and does not modify the underlying OmniLLM. As shown in
Figure~\ref{fig:omnifocus}, OmniFocus has three stages: (1) estimate
query-aware importance for each modality in each chunk, (2) convert these
importance scores into chunk-wise drop ratios under global modality budgets, and
(3) retain tokens using both inter-modal association and intra-modal saliency.

\paragraph{Query relevance-based modality importance scoring.}
Because query-relevant evidence is often concentrated in a few temporal chunks
and may appear in different modalities, OmniFocus estimates modality-specific
relevance for each audio-video chunk with respect to the textual query. We
extract the query tokens from the LLM token embedding layer, as in standard LLM
input processing, so the query tokens are represented in the same input
embedding space as the audio and video tokens consumed by the LLM. Let
$q = \{q_1, q_2, \dots, q_L\} \in \mathbb{R}^{L \times D}$ denote the embedded
query features. We obtain a single query representation by mean-pooling
$\ell_2$-normalized query tokens and normalizing the result:
\begin{equation}
    \bar{q} = \mathrm{Normalize}\!\left(\frac{1}{L} \sum_{l=1}^{L} \hat{q}_l\right), \quad 
    \hat{q}_l = \frac{q_l}{\|q_l\|_2}
\end{equation}
For each temporal chunk $i \in \{1, \dots, T\}$, OmniFocus measures query
relevance independently for video and audio. Let
$x_{v_i} = \{x_{v_i}^1, \dots, x_{v_i}^{n_v^i}\}$ and
$x_{a_i} = \{x_{a_i}^1, \dots, x_{a_i}^{n_a^i}\}$ denote the video and audio
tokens in chunk $i$. The chunk-level relevance scores are
\begin{equation}
    s_v^i = \max_{j}\; \hat{x}_{v_i}^j \cdot \bar{q}, \quad 
    s_a^i = \max_{j}\; \hat{x}_{a_i}^j \cdot \bar{q}
\end{equation}
where $\hat{x}$ denotes an $\ell_2$-normalized token feature. The max operation
captures sparse query-matching evidence: a chunk receives a high score if at
least one token in that modality is strongly related to the query.

We then convert these relevance scores into compression scores via:
\begin{equation}
    c_v^i = 1 - \mathrm{Softmax}(\mathbf{s}_v)_i, \quad 
    c_a^i = 1 - \mathrm{Softmax}(\mathbf{s}_a)_i
\end{equation}
where $\mathbf{s}_v = [s_v^1, \dots, s_v^T]$ and
$\mathbf{s}_a = [s_a^1, \dots, s_a^T]$. Higher relevance therefore produces a
lower compression score, so query-relevant chunks receive a larger token budget.

\paragraph{Chunk-wise local drop ratio calculation.}
To enable content-adaptive compression while keeping the global retained ratio
fixed, OmniFocus maps relevance scores to chunk-wise local drop ratios under
modality-specific budget constraints. Given a global drop ratio
$r_{\mathrm{global}}^{m}$ for modality $m \in \{v, a\}$, OmniFocus assigns a
local drop ratio to each chunk. We set $r_{\min}=0.35$ and $r_{\max}=0.75$ in
all experiments, and linearly map the compression scores to this allowable
range. When $\max_j c_m^j=\min_j c_m^j$, we skip the linear mapping and set
$\tilde{r}_m^i=r_{\mathrm{global}}^m$ for all chunks. Otherwise, we compute
\begin{equation}
    \tilde{r}_m^i = r_{\min} + 
    \frac{c_m^i - \min_j c_m^j}{\max_j c_m^j - \min_j c_m^j} 
    \cdot (r_{\max} - r_{\min}) .
\end{equation}
This mapping gives larger local drop ratios to less query-relevant chunks. To
match the global token budget in the continuous ratio space, we apply a
calibration offset $\delta$ obtained by binary search:
\begin{equation}
    r_m^i = \mathrm{clamp}\!\left(\tilde{r}_m^i + \delta,\; r_{\min},\; r_{\max}\right), 
    \quad \text{s.t.} \quad 
    \sum_{i=1}^{T} r_m^i \cdot n_m^i = r_{\mathrm{global}}^{m} \cdot N_m
    \label{eq:budget_constraint}
\end{equation}
where $n_m^i$ is the number of modality-$m$ tokens in chunk $i$, and
$N_m = \sum_i n_m^i$ is the total number of modality-$m$ tokens. This allocation
keeps the target retention ratio fixed while allowing different temporal chunks
and modalities to receive different local budgets.

In practice, the global modality-specific budgets are calibrated for each
benchmark to ensure fair comparison across compression baselines. We first
follow the OmniZip protocol to determine the benchmark-level compression setting,
and then adjust the global drop ratios accordingly. The chunk-wise allocation of
OmniFocus is still determined by the query-aware relevance scores above.

\paragraph{Dual-score token selection.}
After determining the retained budget for each chunk, OmniFocus uses a
dual-score criterion to preserve tokens that capture both audio-video
correspondence and modality-specific salient evidence. The criterion consists
of two complementary scores: \textit{inter-modal association scores} and
\textit{intra-modal peak scores}. The association score preserves tokens that
are aligned with the other modality, whereas the peak score preserves tokens
that are distinctive within their own modality. We use mean-pooled normalized
tokens as lightweight reference vectors, which keeps the compression procedure
training-free and avoids additional attention computation.

\textbf{Inter-modal association scores.} For each video token in chunk $i$, we
compute its cosine similarity to the mean representation of the audio tokens in
the same chunk:
\begin{equation}
    a_{\mathrm{assoc}, v_i}^j = \hat{x}_{v_i}^j \cdot \bar{x}_{a_i}, \quad 
    \bar{x}_{a_i} = \mathrm{Normalize}\!\left(\frac{1}{n_a^i} \sum_{k=1}^{n_a^i} \hat{x}_{a_i}^k\right) .
\end{equation}
High-association video tokens preserve audio-visual correspondence within the
same temporal window. Audio tokens are scored symmetrically using the video mean
$\bar{x}_{v_i}$:
\begin{equation}
    a_{\mathrm{assoc}, a_i}^j = \hat{x}_{a_i}^j \cdot \bar{x}_{v_i}, \quad 
    \bar{x}_{v_i} = \mathrm{Normalize}\!\left(\frac{1}{n_v^i} \sum_{k=1}^{n_v^i} \hat{x}_{v_i}^k\right) .
\end{equation}

\textbf{Intra-modal peak scores.} Inter-modal association alone can miss
evidence that appears primarily in one modality. We therefore also score tokens
by their distinctiveness within their own modality:
\begin{equation}
    a_{\mathrm{peak}, v_i}^j = 1 - \hat{x}_{v_i}^j \cdot \bar{x}_{v_i}, \quad
    a_{\mathrm{peak}, a_i}^j = 1 - \hat{x}_{a_i}^j \cdot \bar{x}_{a_i} .
\end{equation}
High peak scores indicate tokens that are distinctive relative to the local
modality context, which helps preserve modality-specific evidence that may be
missed by inter-modal association alone.

Given the keep count $k_m^i = n_m^i - \lfloor r_m^i \cdot n_m^i \rfloor$ for
modality $m$ in chunk $i$, we select $\lfloor k_m^i / 2 \rfloor$ tokens with the
highest association scores and the remaining $\lceil k_m^i / 2 \rceil$ tokens
with the highest peak scores, excluding tokens that have already been selected.
We use this equal split as a simple parameter-free default to balance
audio-visual alignment and modality-specific saliency. Finally, OmniFocus masks
out the dropped tokens and feeds the compressed interleaved sequence to the LLM
backbone.

Overall, OmniFocus combines query-aware modality scoring, content-adaptive
budget allocation, and complementary token selection to preserve both
cross-modal correspondence and modality-specific evidence under a fixed token
budget. Because the method is training-free and operates entirely before the
LLM backbone, it can be applied to existing OmniLLMs without parameter updates
while providing a favorable balance between compression efficiency and
audio-visual fidelity.

\section{Experiment}

\subsection{Evaluation setting}

\paragraph{Models.}
We evaluate OmniFocus on Qwen2.5-Omni-3B and Qwen2.5-Omni-7B. Both models use a default temporal window of 2 seconds to organize audio and video tokens. Video frames are encoded by the Qwen2.5-VL vision encoder, while audio signals are encoded by the Whisper-based audio encoder. The visual and audio tokens within each temporal window are then concatenated to form the interleaved multimodal input sequence.

\paragraph{Baselines.}
We compare OmniFocus with three compression baselines. First, we adopt OmniZip~\cite{tao2025omnizip} as the primary omni-modal compression baseline, which estimates event density from audio-side attention and uses it to determine temporal token budgets. Second, following the baseline setting in OmniZip, we include DyCoke~\cite{tao2025dycoke} as a representative vision-language token compression method and use its TTM component for token reduction. Since full-context attention-based compression can cause out-of-memory errors on long audio-visual inputs, we apply the same DyCoke-style token reduction rule to audio and video tokens under the target token budget. Third, we include a random retention baseline, which uniformly samples tokens under the same token retention ratio.

\paragraph{Benchmarks.}
We evaluate on four audio-visual benchmarks. DailyOmni~\cite{zhou2025daily} evaluates audio-visual question answering with an emphasis on cross-modal temporal reasoning in daily-life videos. WorldSense~\cite{hong2025worldsense} assesses real-world omnimodal understanding over synchronized visual, audio, and textual inputs across diverse scenarios. OmniVideoBench~\cite{li2025omnivideobench} focuses on synergistic audio-visual reasoning with modality complementarity and logical consistency. VideoMME~\cite{fu2025video} evaluates video question answering across short, medium, and long videos. We follow the official evaluation protocol of each benchmark and report the corresponding accuracy or benchmark score.

\paragraph{Implementation Details.}
All experiments are conducted on NVIDIA A100 (80\,GB) GPUs, and ``Full Tokens'' denotes the original Qwen2.5-Omni inference without token compression. Unless otherwise specified, OmniFocus uses $r_{\min}=0.35$ and $r_{\max}=0.75$ and reports results under the 35\% and 25\% compression settings. We calibrate benchmark-level audio/video budgets for fair comparison across methods; details are provided in Appendix~\ref{app:implementation_details}.

\begin{table*}[!t]
\centering
\caption{\textbf{DailyOmni results.} Category-wise performance on DailyOmni for Qwen2.5-Omni-7B and Qwen2.5-Omni-3B under 35\% and 25\% token retention. ``Full Tokens'' denotes the uncompressed model. The best compressed result in each column is highlighted in bold.}
\renewcommand{\arraystretch}{1.13}
\setlength{\tabcolsep}{4.2pt}
\small

\begin{adjustbox}{max width=\textwidth}
\begin{tabular}{@{}lcccccccc@{}}
\toprule
Method
& \makecell{Retained\\ Ratio (\%)}
& \makecell{AV Event\\ Alignment ($\uparrow$)}
& Comparative ($\uparrow$)
& \makecell{Context\\ Understanding ($\uparrow$)}
& \makecell{Event\\ Sequence ($\uparrow$)}
& Inference ($\uparrow$)
& Reasoning ($\uparrow$)
& Average ($\uparrow$) \\
\midrule

\multicolumn{9}{@{}l}{Qwen2.5-Omni-7B} \\
\midrule
Full Tokens & 100 & 48.32 & 69.47 & 59.59 & 57.52 & 75.32 & 73.14 & 61.90 \\
\midrule

Random & 35 & 42.86 & \textbf{70.23} & 53.89 & 52.61 & 72.08 & 72.57 & 58.23 \\
DyCoke & 35 & 43.02 & 62.77 & \textbf{57.25} & \textbf{56.64} & 68.48 & 71.84 & 57.57 \\
OmniZip & 35 & 43.28 & 67.94 & 52.85 & 53.59 & \textbf{79.22} & \textbf{73.71} & 59.23 \\
\rowcolor{oursrowbg}
Ours & 35 & \textbf{48.32} & 68.70 & 55.96 & 52.61 & 78.57 & 72.57 & \textbf{60.32} \\
\midrule

Random & 25 & 40.76 & \textbf{69.47} & 54.92 & 53.59 & 75.32 & 68.57 & 57.98 \\
DyCoke & 25 & 38.82 & 55.57 & \textbf{59.35} & \textbf{55.54} & 65.88 & 68.04 & 54.37 \\
OmniZip & 25 & 42.02 & 67.94 & 51.81 & 51.63 & \textbf{75.97} & \textbf{72.57} & 57.73 \\
\rowcolor{oursrowbg}
Ours & 25 & \textbf{46.64} & \textbf{69.47} & 54.92 & 52.29 & 75.32 & \textbf{72.57} & \textbf{59.40} \\
\midrule
\multicolumn{9}{@{}l}{Qwen2.5-Omni-3B} \\
\midrule
Full Tokens & 100 & 51.68 & 69.47 & 53.37 & 53.59 & 74.68 & 70.86 & 60.15 \\
\midrule
Random & 35 & 44.96 & 64.12 & 51.30 & 46.41 & 74.03 & 64.00 & 54.97 \\
DyCoke & 35 & \textbf{48.22} & 57.48 & \textbf{54.49} & \textbf{55.68} & 73.87 & 67.24 & 56.15 \\
OmniZip & 35 & 44.54 & 63.36 & 52.33 & 47.71 & \textbf{75.32} & 66.67 & 55.85 \\
\rowcolor{oursrowbg}
Ours & 35 & 47.90 & \textbf{66.41} & 51.30 & 48.04 & \textbf{75.97} & \textbf{69.14} & \textbf{57.23} \\
\midrule
Random & 25 & 43.70 & \textbf{64.12} & 48.70 & 45.10 & 71.43 & 64.57 & 53.72 \\
DyCoke & 25 & 46.92 & 59.58 & 49.99 & \textbf{53.98} & 71.77 & 64.24 & 54.65 \\
OmniZip & 25 & 46.22 & 61.83 & 49.22 & 48.37 & \textbf{74.68} & 66.09 & 55.52 \\
\rowcolor{oursrowbg}
Ours & 25 & \textbf{47.90} & \textbf{64.12} & \textbf{50.78} & 48.04 & 72.73 & \textbf{66.86} & \textbf{56.14} \\
\bottomrule
\end{tabular}
\end{adjustbox}
\label{tab:dailyomni_comparison}
\end{table*}

\subsection{Main Results}

\begin{table*}[t]
\centering
\caption{\textbf{Results on WorldSense, OmniVideoBench, and VideoMME.} Performance of Qwen2.5-Omni under 35\% and 25\% token retention. ``Full Tokens'' denotes the uncompressed model. The best compressed result in each column is highlighted in bold.}
\renewcommand{\arraystretch}{1.13}
\setlength{\tabcolsep}{4.5pt}
\footnotesize

\begin{adjustbox}{max width=\textwidth}
\begin{tabular}{@{}lccccccc@{}}
\toprule
\multirow{2}{*}{Method}
& \multirow{2}{*}{\makecell{Retained\\ Ratio (\%)}}
& \multirow{2}{*}{WorldSense ($\uparrow$)}
& \multirow{2}{*}{\makecell{OmniVideo ($\uparrow$)\\ Bench}}
& \multicolumn{4}{c}{VideoMME ($\uparrow$)} \\
\cmidrule(lr){5-8}
& & & & Short & Medium & Long & Avg. \\
\midrule

\multicolumn{8}{@{}l}{Qwen2.5-Omni-7B} \\
\midrule
Full Tokens & 100 & 45.81 & 35.31 & 76.00 & 68.33 & 64.89 & 69.74 \\
\midrule
Random & 35 & 44.14 & 33.20 & 72.11 & 65.00 & \textbf{65.67} & 67.59 \\
DyCoke & 35 & 43.28 & 32.30 & 73.01 & 64.33 & 65.33 & 67.99 \\
OmniZip & 35 & 44.55 & 31.49 & 73.67 & 66.00 & 64.33 & 68.00 \\
\rowcolor{oursrowbg}
Ours & 35 & \textbf{44.77} & \textbf{34.10} & \textbf{74.78} & \textbf{68.00} & 65.44 & \textbf{69.41} \\
\midrule
Random & 25 & 43.82 & 32.20 & 71.44 & 65.22 & \textbf{65.33} & 67.33 \\
DyCoke & 25 & 43.25 & 32.19 & 72.01 & 63.33 & 64.33 & 66.99 \\
OmniZip & 25 & 43.69 & 32.30 & 72.89 & 66.33 & 64.67 & 67.96 \\
\rowcolor{oursrowbg}
Ours & 25 & \textbf{44.23} & \textbf{32.70} & \textbf{73.22} & \textbf{66.89} & 64.56 & \textbf{68.22} \\
\midrule
\multicolumn{8}{@{}l}{Qwen2.5-Omni-3B} \\
\midrule
Full Tokens & 100 & 45.21 & 31.09 & 74.11 & 65.56 & 60.11 & 66.59 \\
\midrule
Random & 35 & 43.44 & 32.00 & 67.89 & 62.56 & 59.56 & 63.33 \\
DyCoke & 35 & 43.32 & 31.29 & 67.89 & 62.56 & 59.56 & 63.33 \\
OmniZip & 35 & 43.88 & 32.30 & 70.56 & \textbf{63.67} & 59.89 & 64.70 \\
\rowcolor{oursrowbg}
Ours & 35 & \textbf{44.51} & \textbf{32.40} & \textbf{71.89} & 62.56 & \textbf{60.00} & \textbf{64.81} \\
\midrule
Random & 25 & 42.78 & 31.19 & 67.67 & \textbf{63.22} & 59.78 & 63.56 \\
DyCoke & 25 & 43.25 & 31.19 & 70.01 & 62.10 & 59.78 & 64.00 \\
OmniZip & 25 & 43.06 & 32.10 & 70.56 & \textbf{63.22} & 59.78 & 64.52 \\
\rowcolor{oursrowbg}
Ours & 25 & \textbf{44.10} & \textbf{32.10} & \textbf{71.22} & 62.89 & \textbf{61.00} & \textbf{65.04} \\
\bottomrule
\end{tabular}
\end{adjustbox}
\label{tab:performance_comparison}
\end{table*}

\paragraph{Category-wise performance}

Table~\ref{tab:dailyomni_comparison} provides a category-level breakdown on
DailyOmni. Relative to OmniZip, the most consistent gains appear on
\textit{AV Event Alignment}. On Qwen2.5-Omni-7B, OmniFocus improves over
OmniZip by 5.04 and 4.62 points at 35\% and 25\% retention, respectively; on
Qwen2.5-Omni-3B, the corresponding gains are 3.36 and 1.68 points. The larger
gains on 7B suggest that stronger OmniLLMs can make better use of preserved
aligned audio-visual cues under compression. Other categories are more mixed:
Comparative and Context Understanding show gains in several settings, while
Event Sequence, Inference, and Reasoning remain closer to the baseline. This
pattern indicates that the main benefit of OmniFocus is concentrated on
alignment-sensitive questions rather than uniformly distributed across all
reasoning types.

\paragraph{Results on audio-video benchmarks}

Table~\ref{tab:performance_comparison} summarizes the performance under different
token retention ratios. Across compressed settings, OmniFocus maintains strong
performance on all three audio-video benchmarks, with the clearest consistency
on WorldSense and VideoMME average scores. On WorldSense, OmniFocus achieves
the strongest compressed result in all four model-budget settings,
outperforming OmniZip by 0.22 and 0.54 points on Qwen2.5-Omni-7B at 35\% and
25\% retention, respectively, and by 0.63 and 1.04 points on Qwen2.5-Omni-3B.
This result suggests that query-guided modality-balanced compression is
particularly helpful when the benchmark requires synchronized real-world
perception. On VideoMME, OmniFocus also achieves the best compressed average
score in all four settings, indicating that preserving query-relevant
audio-visual evidence remains beneficial across video question answering
scenarios with different temporal ranges. On OmniVideoBench, the gains are
smaller but still stable: OmniFocus achieves the best compressed score in three
settings and ties the remaining one, showing that the method remains
competitive even when benchmark gains are more modest.

Overall, these results suggest that the main advantage of OmniFocus comes from
preserving cross-modal evidence where compression decisions matter most, rather
than from uniformly improving every benchmark or question type. This pattern is
especially clear on DailyOmni alignment-sensitive categories, where the gains
are larger on Qwen2.5-Omni-7B than on Qwen2.5-Omni-3B. This is consistent with
the view that smaller OmniLLMs may be less effective at exploiting preserved
cross-modal cues under compression, even when the compressor retains them.

Appendix~\ref{app:budget_sensitivity} further analyzes how different global audio-video budget allocations affect performance.

\subsection{Efficiency Analysis}

We further evaluate the accuracy-efficiency trade-off on DailyOmni. In
Table~\ref{tab:efficiency_analysis}, we report DailyOmni accuracy, peak GPU
memory, and relative speedups for prefill time and end-to-end latency. All
speedups are computed against the corresponding full-token inference baseline,
so larger values indicate better efficiency.

\begin{table*}[t]
\centering
\caption{\textbf{Efficiency analysis on DailyOmni.} We report accuracy, peak GPU memory, and relative prefill and end-to-end speedups over the corresponding full-token inference baseline. The left and right subtables report Qwen2.5-Omni-7B and Qwen2.5-Omni-3B results, respectively.}
\renewcommand{\arraystretch}{1.10}
\setlength{\tabcolsep}{3.2pt}
\footnotesize
\vspace{0.45em}
\begin{minipage}[t]{0.495\textwidth}
\centering
\begin{adjustbox}{max width=\linewidth}
\begin{tabular}{@{}lccccc@{}}
\toprule
Method & \makecell{Target\\(\%)} & Acc. & \makecell{GPU\\Mem.\\(GB)} & \makecell{Prefill\\Spd.} & \makecell{E2E\\Spd.} \\
\midrule
Full Tokens & 100 & 61.90 & 19 & 1.00x & 1.00x \\
\midrule
OmniZip & 35 & 58.98 & 18 & 1.31x & 1.24x \\
OmniZip & 25 & 56.98 & 18 & 1.34x & 1.27x \\
\rowcolor{oursrowbg}
\textbf{OmniFocus} & 35 & \textbf{60.32} & 18 & \textbf{1.35x} & \textbf{1.29x} \\
\rowcolor{oursrowbg}
\textbf{OmniFocus} & 25 & \textbf{59.40} & 18 & \textbf{1.38x} & \textbf{1.32x} \\
\bottomrule
\end{tabular}
\end{adjustbox}
\end{minipage}
\hfill
\begin{minipage}[t]{0.495\textwidth}
\centering
\begin{adjustbox}{max width=\linewidth}
\begin{tabular}{@{}lccccc@{}}
\toprule
Method & \makecell{Target\\(\%)} & Acc. & \makecell{GPU\\Mem.\\(GB)} & \makecell{Prefill\\Spd.} & \makecell{E2E\\Spd.} \\
\midrule
Full Tokens & 100 & 60.15 & 11 & 1.00x & 1.00x \\
\midrule
OmniZip & 35 & 55.56 & 10 & 1.15x & 1.15x \\
OmniZip & 25 & 54.72 & 10 & 1.09x & 1.08x \\
\rowcolor{oursrowbg}
\textbf{OmniFocus} & 35 & \textbf{57.23} & 10 & \textbf{1.19x} & \textbf{1.17x} \\
\rowcolor{oursrowbg}
\textbf{OmniFocus} & 25 & \textbf{56.14} & 10 & \textbf{1.20x} & \textbf{1.19x} \\
\bottomrule
\end{tabular}
\end{adjustbox}
\end{minipage}
\label{tab:efficiency_analysis}
\end{table*}

Table~\ref{tab:efficiency_analysis} shows that OmniFocus improves the
accuracy-efficiency trade-off over the compressed baseline. On Qwen2.5-Omni-7B,
OmniFocus improves DailyOmni accuracy by 1.34 and 2.42 points over OmniZip at
35\% and 25\% target retention, respectively, while also providing higher
prefill and end-to-end speedups. On Qwen2.5-Omni-3B, OmniFocus similarly
improves accuracy by 1.67 and 1.42 points and achieves stronger prefill and
end-to-end speedups than OmniZip. Although OmniFocus uses slightly more peak
GPU memory than OmniZip, both compressed methods substantially reduce memory
relative to full-token inference.

\subsection{Ablation Study}

We conduct ablation studies on DailyOmni to examine the main design choices of
OmniFocus. Table~\ref{tab:ablation_qwen7b} reports chunk relevance scoring and
drop-ratio allocation, and score transformation ablations on Qwen2.5-Omni-7B, while
Figure~\ref{fig:ablation_inter_intra} evaluates inter-/intra-modal token
selection for both model sizes. Table~\ref{tab:ablation_modality_guidance}
further compares audio-only, video-only, and modality-balanced chunk scoring.
Unless otherwise specified, the default configuration uses max-similarity
relevance scoring, score-based allocation, softmax score transformation, hybrid
inter+intra token selection, and balanced audio-video relevance estimation.

\begin{table*}[t]
\centering
\begin{minipage}[t]{0.46\textwidth}
\centering
\caption{\textbf{Ablation study on Qwen2.5-Omni-7B.} We report the effects of
chunk relevance scoring, drop ratio allocation, and score transformation on
DailyOmni under 35\% and 25\% token retention. The best result within
each ablation group is highlighted in bold.}
\renewcommand{\arraystretch}{1.12}
\setlength{\tabcolsep}{4.2pt}
\small
\begin{adjustbox}{max width=\linewidth}
\begin{tabular}{@{}llcc@{}}
\toprule
\textbf{Component} & \textbf{Variant} & \textbf{35\%} & \textbf{25\%} \\
\midrule
\multirow{2}{*}{Chunk relevance scoring}
& Average similarity & 59.73 & 58.90 \\
& Max similarity (Ours) & \textbf{60.32} & \textbf{59.40} \\
\midrule
\multirow{3}{*}{Drop ratio allocation}
& Random & 58.81 & 57.31 \\
& Uniform & 59.82 & 59.06 \\
& Score-based (Ours) & \textbf{60.32} & \textbf{59.40} \\
\midrule
\multirow{2}{*}{Score transformation}
& Sigmoid & \textbf{60.32} & 59.31 \\
& Softmax (Ours) & \textbf{60.32} & \textbf{59.40} \\
\bottomrule
\end{tabular}
\end{adjustbox}
\label{tab:ablation_qwen7b}
\end{minipage}
\hfill
\begin{minipage}[t]{0.50\textwidth}
\centering
\vspace{0pt}
\includegraphics[width=0.96\linewidth]{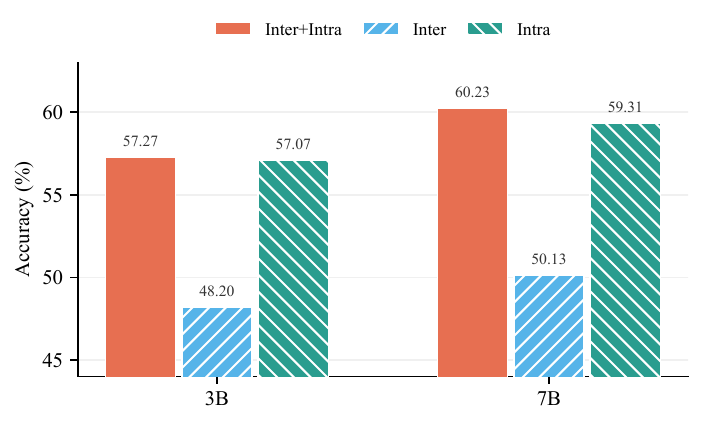}
\captionof{figure}{\textbf{Impact of inter- and intra-modal token selection.}
We compare inter-modal only, intra-modal only, and the hybrid inter+intra
strategy under 35\% token retention for 3B and 7B models.}
\label{fig:ablation_inter_intra}
\end{minipage}
\end{table*}

\textbf{Chunk relevance scoring.} Max-similarity scoring outperforms average
similarity on Qwen2.5-Omni-7B, improving the average score by 0.59 points at
35\% retention and 0.50 points at 25\% retention. This supports the intuition
that query-relevant evidence can be sparse within a temporal chunk and is better
captured by the most responsive token than by mean aggregation.

\textbf{Drop-ratio allocation.} Score-based allocation achieves the best average
scores under both retention ratios, reaching 60.32 at 35\% and 59.40 at 25\%.
Compared with random and uniform allocation, adaptive chunk-wise compression
better preserves informative temporal segments under a fixed token budget.

\textbf{Score transformation.} Softmax and sigmoid produce very similar average
performance, with both reaching 60.32 at 35\% retention. At 25\% retention,
softmax is slightly better than sigmoid (59.40 vs. 59.31), so we use softmax as
the default transformation.

\textbf{Inter- and intra-modal token selection.} The hybrid inter+intra strategy
achieves the best result for both model sizes in Figure~\ref{fig:ablation_inter_intra}.
Inter-modal association alone loses important modality-specific evidence, while
intra-modal selection recovers most of the performance; combining both cues
therefore provides a better balance between audio-visual correspondence and
within-modality saliency. A sensitivity analysis of the inter/intra keep-count
split is provided in Appendix~\ref{app:inter_intra_ratio}.

\textbf{Modality-guided chunk scoring.} Balanced scoring achieves the highest
average score in all four model-budget settings, indicating that query-relevant
chunk importance is more reliably estimated when audio and video are considered
jointly rather than in isolation. This suggests that neither modality alone can
consistently capture the evidence distribution required for compressing the
other modality. As detailed in Appendix~\ref{app:modality_guidance_detail},
single-modality variants can still perform better on some individual
categories, but their overall performance is less stable than that of
modality-balanced scoring.

\begin{table*}[t]
\centering
\caption{\textbf{Average ablation of modality-guided chunk scoring on DailyOmni.}
We compare audio-only, video-only, and modality-balanced chunk scoring under
35\% and 25\% token retention. The left and right subtables report
Qwen2.5-Omni-7B and Qwen2.5-Omni-3B results, respectively. Detailed
category-wise results are provided in Appendix~\ref{app:modality_guidance_detail}.}
\renewcommand{\arraystretch}{1.12}
\setlength{\tabcolsep}{5.5pt}
\footnotesize
\vspace{0.35em}
\begin{minipage}[t]{0.48\textwidth}
\centering
\begin{adjustbox}{max width=\linewidth}
\begin{tabular}{@{}lcc@{}}
\toprule
Scoring Strategy & \makecell{Retained\\ Ratio (\%)} & Average ($\uparrow$) \\
\midrule
Audio-only & 35 & 60.14 \\
Video-only & 35 & 60.07 \\
\rowcolor{oursrowbg}
Balanced (Ours) & 35 & \textbf{60.32} \\
\midrule
Audio-only & 25 & 59.05 \\
Video-only & 25 & 58.65 \\
\rowcolor{oursrowbg}
Balanced (Ours) & 25 & \textbf{59.40} \\
\bottomrule
\end{tabular}
\end{adjustbox}
\end{minipage}
\hfill
\begin{minipage}[t]{0.48\textwidth}
\centering
\begin{adjustbox}{max width=\linewidth}
\begin{tabular}{@{}lcc@{}}
\toprule
Scoring Strategy & \makecell{Retained\\ Ratio (\%)} & Average ($\uparrow$) \\
\midrule
Audio-only & 35 & 56.06 \\
Video-only & 35 & 56.98 \\
\rowcolor{oursrowbg}
Balanced (Ours) & 35 & \textbf{57.23} \\
\midrule
Audio-only & 25 & 55.72 \\
Video-only & 25 & 56.10 \\
\rowcolor{oursrowbg}
Balanced (Ours) & 25 & \textbf{56.14} \\
\bottomrule
\end{tabular}
\end{adjustbox}
\end{minipage}
\label{tab:ablation_modality_guidance}
\end{table*}

\section{Conclusion}

We introduced OmniFocus, a training-free query-guided token compression method
for reducing the inference cost of omni-modal large language models on long
audio-visual inputs.
By estimating query relevance separately for audio and video and selecting
tokens with both inter-modal association and intra-modal peak evidence,
OmniFocus reduces the modality bias of unimodal-guided compression.
Experiments on Qwen2.5-Omni-3B and Qwen2.5-Omni-7B show that OmniFocus achieves
a favorable accuracy-efficiency trade-off across audio-visual benchmarks, while
ablations validate the contribution of its core design choices.

\section{Limitations}

OmniFocus relies on query-token similarity in the frozen input embedding space
and is currently evaluated mainly on the Qwen2.5-Omni family. Detailed
limitations are discussed in Appendix~\ref{app:limitations}.

\bibliographystyle{plainnat}
\bibliography{references}


\clearpage
\appendix
\input{appendix.tex}


\clearpage
\input{checklist.tex}

\end{document}

%% file: appendix.tex
\section{Implementation Details}
\label{app:implementation_details}

This section provides additional implementation details omitted from the main text, with an emphasis on the budget calibration protocol and default hyperparameter settings.

OmniFocus is training-free: all token compression operations are performed before the LLM backbone, and no model parameters are updated during evaluation. We use the same Qwen2.5-Omni input organization as the full-token setting, where audio and video tokens are grouped by the default 2-second temporal window before compression. ``Full Tokens'' denotes the original Qwen2.5-Omni inference path without any token dropping.

For fair comparison, we calibrate compression budgets separately for each benchmark. We first follow the OmniZip protocol to determine the benchmark-level compression setting and then adjust the global modality-specific drop ratios for audio and video accordingly. This benchmark-level calibration is used consistently across methods.

Unless otherwise specified, OmniFocus uses $r_{\min}=0.35$ and $r_{\max}=0.75$ as the local drop-ratio range. Within each chunk and modality, the retained tokens are selected with a 1:1 split between inter-modal association scores and intra-modal peak scores.

\section{Modality-Type Analysis Protocol}
\label{app:modality_type_analysis}

To analyze whether compression errors are concentrated on questions requiring specific modality information, we use Qwen3-32B to classify each multiple-choice video question into one of three modality types: \textit{video}, \textit{audio}, or \textit{video-audio}.

The classifier only receives the question and answer options, without access to the video, audio, model prediction, or ground-truth answer. Therefore, this analysis does not leak evaluation labels and is independent of the prediction results of the evaluated OmniLLMs.

We use the following prompt for question-type classification and require the model to return exactly one label:

\begin{quote}
\small
\ttfamily
You are classifying what modality information is required to answer a multiple-choice video question.\\
You must choose exactly one label from: video, audio, video-audio.\\[0.4em]
Definitions:\\
- video: the question can be answered from visual information alone.\\
- audio: the question can be answered from audio information alone.\\
- video-audio: the question requires both visual and audio information.\\[0.4em]
Return only one label: video or audio or video-audio.\\[0.4em]
Question:\\
\{question\}\\[0.4em]
Options:\\
\{options\_text\}
\end{quote}

After obtaining the modality-type labels, we group evaluation examples by their predicted modality requirement and compute the error distribution of each compression method within these groups. This allows us to examine whether a compression strategy disproportionately harms visual evidence, audio evidence, or joint audio-visual evidence.

\section{Additional Efficiency Analysis}
\label{app:duration_efficiency}

We further analyze efficiency on WorldSense by grouping QA examples according
to the duration of their source video. Each example is assigned to a
100-second bucket based on the actual video length measured by
\texttt{ffprobe}, and we report results for the 0--600s range.

\begin{figure}[t]
\centering
\includegraphics[width=\linewidth]{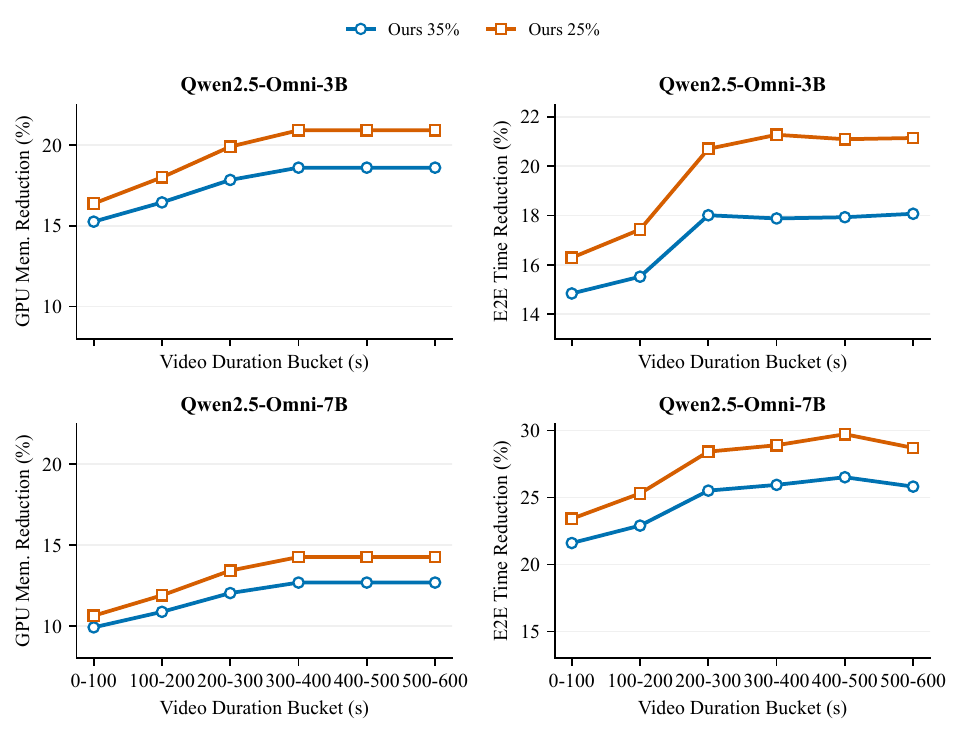}
\caption{\textbf{Efficiency gains by video duration on WorldSense.} We group QA examples into 100-second video-duration buckets and report GPU memory reduction and end-to-end time reduction relative to the same-model full-token baseline. Longer videos generally yield larger relative savings, and the stronger compression setting (25\%) provides the largest gains.}
\label{fig:duration_bucket_efficiency}
\end{figure}

Figure~\ref{fig:duration_bucket_efficiency} shows that the relative efficiency
gains of OmniFocus on WorldSense generally increase with video duration and
then become more stable after roughly the 300s bucket. On Qwen2.5-Omni-3B, GPU
memory reduction grows from 15.25\% to 18.59\% at 35\% retention and from
16.37\% to 20.90\% at 25\% retention, while the corresponding end-to-end time
reduction rises from 14.84\% to 18.07\% and from 16.29\% to 21.14\%. On
Qwen2.5-Omni-7B, the GPU memory reduction is smaller in magnitude but still
increases with duration, reaching 12.68\% at 35\% retention and 14.26\% at
25\% retention, whereas the end-to-end time reduction is more pronounced,
reaching 25.81\% and 28.70\% in the longest bucket. These results suggest that
compression becomes increasingly valuable on longer WorldSense videos,
especially for stronger compression settings and larger models in end-to-end
latency.

\section{Additional Ablation Results}
\label{app:additional_ablation}

\subsection{Sensitivity to Modality-Specific Budgets}
\label{app:budget_sensitivity}

\begin{figure}[t]
\centering
\includegraphics[width=\linewidth,height=0.20\textheight,keepaspectratio]{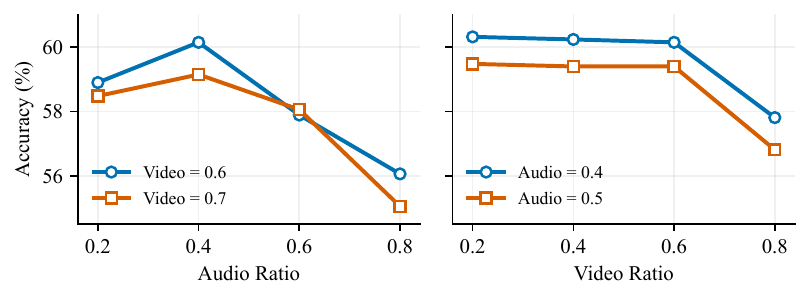}
\caption{\textbf{Sensitivity to modality-specific retention ratios.} We vary the audio retention ratio while fixing the video ratio (left), and vary the video retention ratio while fixing the audio ratio (right). OmniFocus is robust around moderate modality budgets, while overly aggressive compression of either modality degrades DailyOmni performance.}
\label{fig:retention_ratio}
\end{figure}

We further analyze the sensitivity of OmniFocus to different modality-specific retention budgets on DailyOmni. Figure~\ref{fig:retention_ratio} varies the audio retention ratio with a fixed video ratio and varies the video retention ratio with a fixed audio ratio.

Figure~\ref{fig:retention_ratio} shows that moderate audio-video budget choices preserve stable performance, whereas overly aggressive compression of either modality can remove necessary modality-specific evidence. This supports the use of calibrated global modality-specific budgets instead of a single fixed unimodal compression rule.

\subsection{Detailed Modality-Guided Chunk Scoring Results}
\label{app:modality_guidance_detail}

Table~\ref{tab:app_modality_guidance_detail} provides the category-wise results omitted from the average-only summary in Table~\ref{tab:ablation_modality_guidance}.

\begin{table*}[t]
\centering
\caption{\textbf{Detailed ablation of modality-guided chunk scoring on DailyOmni.} We compare audio-only, video-only, and modality-balanced chunk scoring under 35\% and 25\% token retention. The best result within each model-budget group is highlighted in bold.}
\renewcommand{\arraystretch}{1.13}
\setlength{\tabcolsep}{4.2pt}
\small
\begin{adjustbox}{max width=\textwidth}
\begin{tabular}{@{}lcccccccc@{}}
\toprule
Scoring Strategy
& \makecell{Retained\\ Ratio (\%)}
& \makecell{AV Event\\ Alignment ($\uparrow$)}
& Comparative ($\uparrow$)
& \makecell{Context\\ Understanding ($\uparrow$)}
& \makecell{Event\\ Sequence ($\uparrow$)}
& Inference ($\uparrow$)
& Reasoning ($\uparrow$)
& Average ($\uparrow$) \\
\midrule
\multicolumn{9}{@{}l}{Qwen2.5-Omni-7B} \\
\midrule
Audio-only & 35 & \textbf{49.16} & 67.94 & 55.44 & \textbf{54.90} & 75.32 & \textbf{74.29} & 60.14 \\
Video-only & 35 & 47.90 & \textbf{68.70} & 53.89 & 52.94 & \textbf{78.57} & 73.14 & 60.07 \\
\rowcolor{oursrowbg}
Balanced (Ours) & 35 & 48.32 & \textbf{68.70} & \textbf{55.96} & 52.61 & \textbf{78.57} & 72.57 & \textbf{60.32} \\
\midrule
Audio-only & 25 & 44.96 & 67.94 & \textbf{56.99} & \textbf{54.58} & 74.03 & 72.57 & 59.05 \\
Video-only & 25 & 46.22 & 67.94 & 52.33 & 50.98 & \textbf{76.62} & \textbf{73.14} & 58.65 \\
\rowcolor{oursrowbg}
Balanced (Ours) & 25 & \textbf{46.64} & \textbf{69.47} & 54.92 & 52.29 & 75.32 & 72.57 & \textbf{59.40} \\
\midrule
\multicolumn{9}{@{}l}{Qwen2.5-Omni-3B} \\
\midrule
Audio-only & 35 & 46.64 & \textbf{66.41} & \textbf{51.30} & 47.39 & 73.38 & 66.29 & 56.06 \\
Video-only & 35 & \textbf{48.32} & 64.89 & \textbf{51.30} & \textbf{49.67} & 74.03 & 66.86 & 56.98 \\
\rowcolor{oursrowbg}
Balanced (Ours) & 35 & 47.90 & \textbf{66.41} & \textbf{51.30} & 48.04 & \textbf{75.97} & \textbf{69.14} & \textbf{57.23} \\
\midrule
Audio-only & 25 & 46.64 & \textbf{64.12} & 49.22 & 48.37 & 72.73 & 66.86 & 55.72 \\
Video-only & 25 & 47.06 & \textbf{64.12} & \textbf{51.30} & \textbf{49.67} & \textbf{73.38} & \textbf{67.43} & 56.10 \\
\rowcolor{oursrowbg}
Balanced (Ours) & 25 & \textbf{47.90} & \textbf{64.12} & 50.78 & 48.04 & 72.73 & 66.86 & \textbf{56.14} \\
\bottomrule
\end{tabular}
\end{adjustbox}
\label{tab:app_modality_guidance_detail}
\end{table*}

\subsection{Sensitivity to Inter/Intra Token-Selection Split}
\label{app:inter_intra_ratio}

We further analyze the keep-count split between inter-modal association tokens
and intra-modal peak tokens.
This ablation only changes the inter/intra split while keeping the other
DailyOmni settings unchanged.

\begin{figure}[t]
\centering
\includegraphics[width=\linewidth]{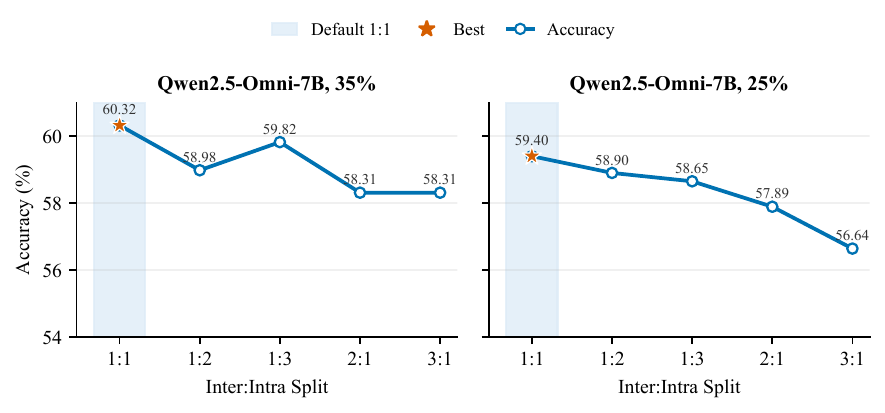}
\caption{\textbf{Sensitivity to the inter/intra token-selection split on DailyOmni for Qwen2.5-Omni-7B.}
We report average accuracy under different keep-count splits between inter-modal
association scores and intra-modal peak scores at 35\% and 25\% token
retention. The default 1:1 split is highlighted, and the best point in each
retention setting is marked.}
\label{fig:app_inter_intra_ratio}
\end{figure}

Figure~\ref{fig:app_inter_intra_ratio} shows that intra-heavy splits generally
outperform inter-heavy splits on Qwen2.5-Omni-7B, suggesting that
modality-internal peak evidence is particularly useful for token selection on
DailyOmni. The default 1:1 split is best at 35\% retention and remains close to
the best configuration at 25\% retention. We therefore keep the symmetric 1:1
split as the default.

\subsection{Robustness of Reference Vector Estimation}
\label{app:reference_vector_ablation}

Table~\ref{tab:app_reference_vector_ablation} further tests whether the reference vectors used by inter-modal and intra-modal token selection require more robust centroid estimation. This ablation only changes the reference vector strategy, while the chunk-level query relevance score still uses max similarity.

\begin{table*}[t]
\centering
\caption{\textbf{Robust reference vector ablation on DailyOmni.} We compare mean, trimmed-mean, and query-weighted reference vectors for inter-/intra-modal token selection on Qwen2.5-Omni-7B. Only the reference vector estimation is changed; chunk-level query relevance still uses max similarity. The best average score within each retention group is highlighted in bold.}
\renewcommand{\arraystretch}{1.13}
\setlength{\tabcolsep}{4.2pt}
\small
\begin{adjustbox}{max width=\textwidth}
\begin{tabular}{@{}lcccccccc@{}}
\toprule
\makecell{Reference\\Strategy}
& \makecell{Retained\\Ratio (\%)}
& \makecell{AV Event\\Alignment ($\uparrow$)}
& Comparative ($\uparrow$)
& \makecell{Context\\Understanding ($\uparrow$)}
& \makecell{Event\\Sequence ($\uparrow$)}
& Inference ($\uparrow$)
& Reasoning ($\uparrow$)
& Average ($\uparrow$) \\
\midrule
\multicolumn{9}{@{}l}{Qwen2.5-Omni-7B} \\
\midrule
\rowcolor{oursrowbg}
Mean (Ours) & 35 & 48.32 & 68.70 & 55.96 & 52.61 & 78.57 & 72.57 & \textbf{60.32} \\
Trimmed Mean & 35 & 44.96 & 68.70 & 54.40 & 52.61 & 77.27 & 73.14 & 59.31 \\
Query Weighted & 35 & 46.64 & 67.18 & 53.37 & 51.96 & 73.38 & 72.00 & 58.48 \\
\midrule
\rowcolor{oursrowbg}
Mean (Ours) & 25 & 46.64 & 69.47 & 54.92 & 52.29 & 75.32 & 72.57 & \textbf{59.40} \\
Trimmed Mean & 25 & 44.12 & 68.70 & 53.89 & 51.31 & 77.27 & 71.43 & 58.48 \\
Query Weighted & 25 & 45.80 & 67.18 & 52.85 & 50.65 & 70.13 & 71.43 & 57.39 \\
\bottomrule
\end{tabular}
\end{adjustbox}
\label{tab:app_reference_vector_ablation}
\end{table*}

Table~\ref{tab:app_reference_vector_ablation} shows that the default mean centroid achieves the highest average score under both retention ratios. Trimmed-mean and query-weighted centroids do not improve robustness in this setting, suggesting that the simple mean reference vector is already sufficiently stable for the normalized token features used by OmniFocus. This also keeps the token selection module lightweight and avoids introducing an additional robust-estimation hyperparameter beyond the compression budget.

\subsection{Sensitivity to Query Representation}
\label{app:query_representation}

We further test whether chunk-level query relevance scoring is sensitive to how
the textual query is aggregated into a single representation. Specifically, we
compare the default setting, which averages normalized query-token embeddings,
with mean pooling over the raw query-token embeddings. Table~\ref{tab:app_query_representation}
shows that the two variants are overall comparable, while mean-normalized query
representations are more consistent across model sizes and retention settings.

\begin{table}[t]
\centering
\caption{\textbf{Sensitivity to query representation on DailyOmni.} We compare the default mean-normalized query representation with mean pooling over raw query-token embeddings. The default setting is shaded, and the best score within each model/retention pair is highlighted in bold.}
\renewcommand{\arraystretch}{1.12}
\setlength{\tabcolsep}{4.4pt}
\small
\begin{tabular}{@{}lccc@{}}
\toprule
\makecell{Query\\Representation}
& \makecell{Retained\\Ratio (\%)}
& \makecell{Qwen2.5-\\Omni-7B}
& \makecell{Qwen2.5-\\Omni-3B} \\
\midrule
\rowcolor{oursrowbg}
Mean normalized (default) & 35 & \textbf{60.32} & \textbf{57.23} \\
Mean raw & 35 & 59.82 & 57.14 \\
\midrule
\rowcolor{oursrowbg}
Mean normalized (default) & 25 & \textbf{59.40} & \textbf{56.14} \\
Mean raw & 25 & 59.00 & 56.06 \\
\bottomrule
\end{tabular}
\label{tab:app_query_representation}
\end{table}

The default mean-normalized query representation is better in three of the four
settings and remains close to the best variant in the remaining case. We
therefore keep it as the default because it provides slightly more consistent
performance while staying well aligned with the cosine-similarity scoring used
throughout OmniFocus.

\subsection{Sensitivity to Local Drop-Ratio Range}
\label{app:r_window}

We further analyze the sensitivity of OmniFocus to the local drop-ratio range $[r_{\min}, r_{\max}]$ on DailyOmni. Table~\ref{tab:app_r_window} compares the default range $[0.35, 0.75]$ with a lower range $[0.25, 0.65]$ and a higher range $[0.45, 0.85]$.

\begin{table*}[t]
\centering
\caption{\textbf{Sensitivity to the local drop-ratio range on DailyOmni.} We report category-wise results for Qwen2.5-Omni-7B and Qwen2.5-Omni-3B. The default range used in our main experiments is shaded, and the best average score within each model group is highlighted in bold.}
\renewcommand{\arraystretch}{1.13}
\setlength{\tabcolsep}{4.2pt}
\small
\begin{adjustbox}{max width=\textwidth}
\begin{tabular}{@{}lccccccc@{}}
\toprule
\makecell{Local Drop-Ratio\\Range}
& \makecell{AV Event\\Alignment ($\uparrow$)}
& Comparative ($\uparrow$)
& \makecell{Context\\Understanding ($\uparrow$)}
& \makecell{Event\\Sequence ($\uparrow$)}
& Inference ($\uparrow$)
& Reasoning ($\uparrow$)
& Average ($\uparrow$) \\
\midrule
\multicolumn{8}{@{}l}{Qwen2.5-Omni-7B} \\
\midrule
\rowcolor{oursrowbg}
$[0.35, 0.75]$ & 48.32 & 68.70 & 55.96 & 52.61 & 78.57 & 72.57 & \textbf{60.32} \\
$[0.25, 0.65]$ & 47.48 & 69.47 & 54.92 & 53.27 & 76.62 & 71.43 & 59.82 \\
$[0.45, 0.85]$ & 47.90 & 66.41 & 53.37 & 52.29 & 74.68 & 72.00 & 58.90 \\
\midrule
\multicolumn{8}{@{}l}{Qwen2.5-Omni-3B} \\
\midrule
\rowcolor{oursrowbg}
$[0.35, 0.75]$ & 47.90 & 66.41 & 51.30 & 48.04 & 75.97 & 69.14 & 57.23 \\
$[0.25, 0.65]$ & 50.42 & 64.89 & 50.26 & 50.65 & 75.97 & 68.00 & \textbf{57.89} \\
$[0.45, 0.85]$ & 48.32 & 64.12 & 51.30 & 49.67 & 74.03 & 66.29 & 56.81 \\
\bottomrule
\end{tabular}
\end{adjustbox}
\label{tab:app_r_window}
\end{table*}

Table~\ref{tab:app_r_window} shows that an overly high drop-ratio range degrades the average score for both model sizes, suggesting that overly aggressive local compression can remove useful audio-visual evidence. The default range $[0.35, 0.75]$ performs best on Qwen2.5-Omni-7B, while the lower range $[0.25, 0.65]$ performs best on Qwen2.5-Omni-3B. We use $[0.35, 0.75]$ as the default setting because it provides the strongest 7B performance and avoids the clear degradation observed with the higher range.

\section{Limitations}
\label{app:limitations}

OmniFocus relies on similarity in the frozen input embedding space to perform query-guided compression. This design keeps the method training-free and easy to apply, but the resulting relevance estimates may be less precise for fine-grained events or semantically ambiguous questions where the initial token embeddings do not fully capture the intent of the query.

Our experiments are mainly conducted on the Qwen2.5-Omni model family. Although OmniFocus does not depend on model-specific parameter updates, its generality should be further evaluated on more OmniLLM architectures and under more challenging audio-visual conditions, such as noisy audio, visual occlusion, rapid scene changes, or temporal misalignment between audio and video.

This work uses benchmark-level budget calibration to ensure comparable token retention across compression methods. While this protocol supports fair evaluation, it does not dynamically adapt the global audio and video budgets to the difficulty or modality requirements of each individual example. A promising extension is to develop instance-adaptive budget allocation while maintaining a controlled retained-token ratio.

The training-free design makes OmniFocus lightweight and broadly applicable, but it also limits the ability of the compression policy to optimize directly for downstream task objectives. Future work could combine query-aware compression with lightweight adaptation, calibration data, or internal attention signals from the OmniLLM to further improve task-specific retention.

%% file: checklist.tex
\section*{NeurIPS Paper Checklist}

\begin{enumerate}

\item {\bf Claims}
    \item[] Question: Do the main claims made in the abstract and introduction accurately reflect the paper's contributions and scope?
    \item[] Answer: \answerYes{}
    \item[] Justification: The abstract and introduction state the scope of OmniFocus as a training-free query-guided token compression method for OmniLLMs. The claims have been revised to reflect the experimental results accurately, including competitive rather than uniformly superior performance on all benchmarks.

\item {\bf Limitations}
    \item[] Question: Does the paper discuss the limitations of the work performed by the authors?
    \item[] Answer: \answerYes{}
    \item[] Justification: The paper includes a brief dedicated Limitations section in the main text, and a more detailed discussion of limitations is provided in Appendix~\ref{app:limitations}.

\item {\bf Theory assumptions and proofs}
    \item[] Question: For each theoretical result, does the paper provide the full set of assumptions and a complete (and correct) proof?
    \item[] Answer: \answerNA{}
    \item[] Justification: The paper does not present theoretical theorems or formal proof-based results. The proposed method is evaluated empirically.

    \item {\bf Experimental result reproducibility}
    \item[] Question: Does the paper fully disclose all the information needed to reproduce the main experimental results of the paper to the extent that it affects the main claims and/or conclusions of the paper (regardless of whether the code and data are provided or not)?
    \item[] Answer: \answerYes{}
    \item[] Justification: The paper describes the model family, benchmarks, baselines, token retention settings, budget calibration protocol, local drop-ratio range, and evaluation metrics. The method is training-free, and its scoring, allocation, and token selection procedures are specified in the Method section.

\item {\bf Open access to data and code}
    \item[] Question: Does the paper provide open access to the data and code, with sufficient instructions to faithfully reproduce the main experimental results, as described in supplemental material?
    \item[] Answer: \answerNo{}
    \item[] Justification: The current submission does not provide an open-source code release or detailed reproduction scripts. The experiments use publicly described benchmarks and open model families, and the paper provides methodological and evaluation details for reproducibility.

\item {\bf Experimental setting/details}
    \item[] Question: Does the paper specify all the training and test details (e.g., data splits, hyperparameters, how they were chosen, type of optimizer) necessary to understand the results?
    \item[] Answer: \answerYes{}
    \item[] Justification: The paper specifies the evaluated Qwen2.5-Omni model variants, benchmarks, baselines, 35\% and 25\% token retention settings, global budget calibration protocol, local drop-ratio range, and hardware setup. Since OmniFocus is training-free, no optimizer or training hyperparameters are required.

\item {\bf Experiment statistical significance}
    \item[] Question: Does the paper report error bars suitably and correctly defined or other appropriate information about the statistical significance of the experiments?
    \item[] Answer: \answerNo{}
    \item[] Justification: The current experiments report single-run benchmark scores without error bars, confidence intervals, or statistical significance tests. This is mainly due to the high cost of repeated evaluation on long audio-visual benchmarks and large OmniLLM backbones.

\item {\bf Experiments compute resources}
    \item[] Question: For each experiment, does the paper provide sufficient information on the computer resources (type of compute workers, memory, time of execution) needed to reproduce the experiments?
    \item[] Answer: \answerNo{}
    \item[] Justification: The paper reports that experiments are conducted on NVIDIA A100 80GB GPUs and provides runtime and memory profiling for DailyOmni. However, it does not fully report the total compute time required for every benchmark and ablation experiment.
    
\item {\bf Code of ethics}
    \item[] Question: Does the research conducted in the paper conform, in every respect, with the NeurIPS Code of Ethics \url{https://neurips.cc/public/EthicsGuidelines}?
    \item[] Answer: \answerYes{}
    \item[] Justification: The work focuses on training-free token compression for existing OmniLLMs and evaluates on existing public benchmarks. We are not aware of any deviation from the NeurIPS Code of Ethics.

\item {\bf Broader impacts}
    \item[] Question: Does the paper discuss both potential positive societal impacts and negative societal impacts of the work performed?
    \item[] Answer: \answerNo{}
    \item[] Justification: The current paper does not include a dedicated broader impacts discussion. The work may have positive impacts by reducing the computational cost of long audio-visual inference, but more efficient OmniLLM inference could also lower the cost of misuse in surveillance or multimedia analysis applications.
    
\item {\bf Safeguards}
    \item[] Question: Does the paper describe safeguards that have been put in place for responsible release of data or models that have a high risk for misuse (e.g., pre-trained language models, image generators, or scraped datasets)?
    \item[] Answer: \answerNA{}
    \item[] Justification: The paper does not release a new pretrained model, scraped dataset, or other high-risk asset. It proposes a training-free compression method evaluated on existing models and benchmarks.

\item {\bf Licenses for existing assets}
    \item[] Question: Are the creators or original owners of assets (e.g., code, data, models), used in the paper, properly credited and are the license and terms of use explicitly mentioned and properly respected?
    \item[] Answer: \answerNo{}
    \item[] Justification: The paper cites the existing models, baselines, and benchmarks used in the experiments, but the current draft does not explicitly list their licenses or terms of use.

\item {\bf New assets}
    \item[] Question: Are new assets introduced in the paper well documented and is the documentation provided alongside the assets?
    \item[] Answer: \answerNA{}
    \item[] Justification: The paper does not introduce or release a new dataset, model checkpoint, or benchmark asset. The contribution is a training-free token compression method.

\item {\bf Crowdsourcing and research with human subjects}
    \item[] Question: For crowdsourcing experiments and research with human subjects, does the paper include the full text of instructions given to participants and screenshots, if applicable, as well as details about compensation (if any)? 
    \item[] Answer: \answerNA{}
    \item[] Justification: The paper does not involve crowdsourcing experiments or research with human subjects.

\item {\bf Institutional review board (IRB) approvals or equivalent for research with human subjects}
    \item[] Question: Does the paper describe potential risks incurred by study participants, whether such risks were disclosed to the subjects, and whether Institutional Review Board (IRB) approvals (or an equivalent approval/review based on the requirements of your country or institution) were obtained?
    \item[] Answer: \answerNA{}
    \item[] Justification: The paper does not involve crowdsourcing experiments or human-subject research, so IRB approval is not applicable.

\item {\bf Declaration of LLM usage}
    \item[] Question: Does the paper describe the usage of LLMs if it is an important, original, or non-standard component of the core methods in this research? Note that if the LLM is used only for writing, editing, or formatting purposes and does \emph{not} impact the core methodology, scientific rigor, or originality of the research, declaration is not required.
    \item[] Answer: \answerYes{}
    \item[] Justification: The paper explicitly studies token compression for OmniLLMs and evaluates the method on Qwen2.5-Omni-3B and Qwen2.5-Omni-7B. The role of the LLM backbone and the placement of OmniFocus before the LLM backbone are described in the Method and Experiment sections.

\end{enumerate}